\documentclass[letterpaper, 10 pt, conference]{ieeeconf}  
\IEEEoverridecommandlockouts                              
\pdfoutput=1

\usepackage{graphicx}
\usepackage{float}
\usepackage[normalem]{ulem}
\usepackage{amsmath}
\usepackage[noadjust]{cite}
\usepackage{siunitx}

\makeatletter
\let\NAT@parse\undefined
\makeatother
\usepackage[colorlinks,bookmarksopen,bookmarksnumbered,citecolor=red,urlcolor=red]{hyperref}
\usepackage[capitalise]{cleveref}

\title{\LARGE \bf
GelBelt: A Vision-based Tactile Sensor for Continuous Sensing of Large Surfaces 
}

\begin{document}
\author{Mohammad Amin Mirzaee$^{1}$, Hung-Jui Huang$^{2}$, and Wenzhen Yuan$^{1}$
\thanks{$^{1}$Mohammad Amin Mirzaee and Wenzhen Yuan are with University of Illinois at Urbana-Champaign, Champaign, IL, USA 
        {\tt\small \{mirzaee2,yuanwz\}@illinois.edu}}%
\thanks{$^{2}$Hung-Jui Huang is with Carnegie Mellon University, Pittsburgh, PA, USA
        {\tt\small \{hungjuih\}@andrew.cmu.edu}%
}
}
\maketitle

\thispagestyle{empty}
\pagestyle{empty}

\begin{abstract}
Scanning large-scale surfaces is widely demanded in surface reconstruction applications and detecting defects in industries' quality control and maintenance stages. Traditional vision-based tactile sensors have shown promising performance in high-resolution shape reconstruction while suffering limitations such as small sensing areas or susceptibility to damage when slid across surfaces, making them unsuitable for continuous sensing on large surfaces. To address these shortcomings, we introduce a novel vision-based tactile sensor designed for continuous surface sensing applications. Our design uses an elastomeric belt and two wheels to continuously scan the target surface. The proposed sensor showed promising results in both shape reconstruction and surface fusion, indicating its applicability. The dot product of the estimated and reference surface normal map is reported over the sensing area and for different scanning speeds. Results indicate that the proposed sensor can rapidly scan large-scale surfaces with high accuracy at speeds up to 45 mm/s.
\end{abstract}

\section{Introduction}
Automated surface inspection during quality control 
of manufacturing processes has been demanded since the late 20th century to prevent damage to production \cite{dupont1997optimization}. Subsequently, as industries continue to grow and adopt automation, the demand for reliable solutions has intensified \cite{zheng2021recent,wang2020smart}. Whether for quality control, maintenance, or safety purposes, accurate surface assessment plays a crucial role in ensuring operational efficiency and product quality \cite{wang2020smart}.

Many industries involving large-scale manufacturing machinery and the production of large metal components, like aircraft parts, face challenges such as vibration and debris from foreign objects, high temperatures, friction, and corrosion in the production and maintenance stage of the components \cite{agarwal2023robotic}. Such factors can contribute to fatigue and failure of components, adversely affecting system performance and resulting in irreparable damages \cite{dastgerdi2022influence}. Therefore, each industry has specific maintenance requirements, such as surface inspections, to guarantee safe operations. Accordingly, the need for an automated system that can provide continuous, real-time feedback on the condition of surfaces, ranging from smooth to irregular and textured shapes, grows. Additionally, the proposed system should be sensitive enough to catch detailed information such as tiny defects on the surface.

\begin{figure}
    \centering
    \includegraphics[width=\linewidth]{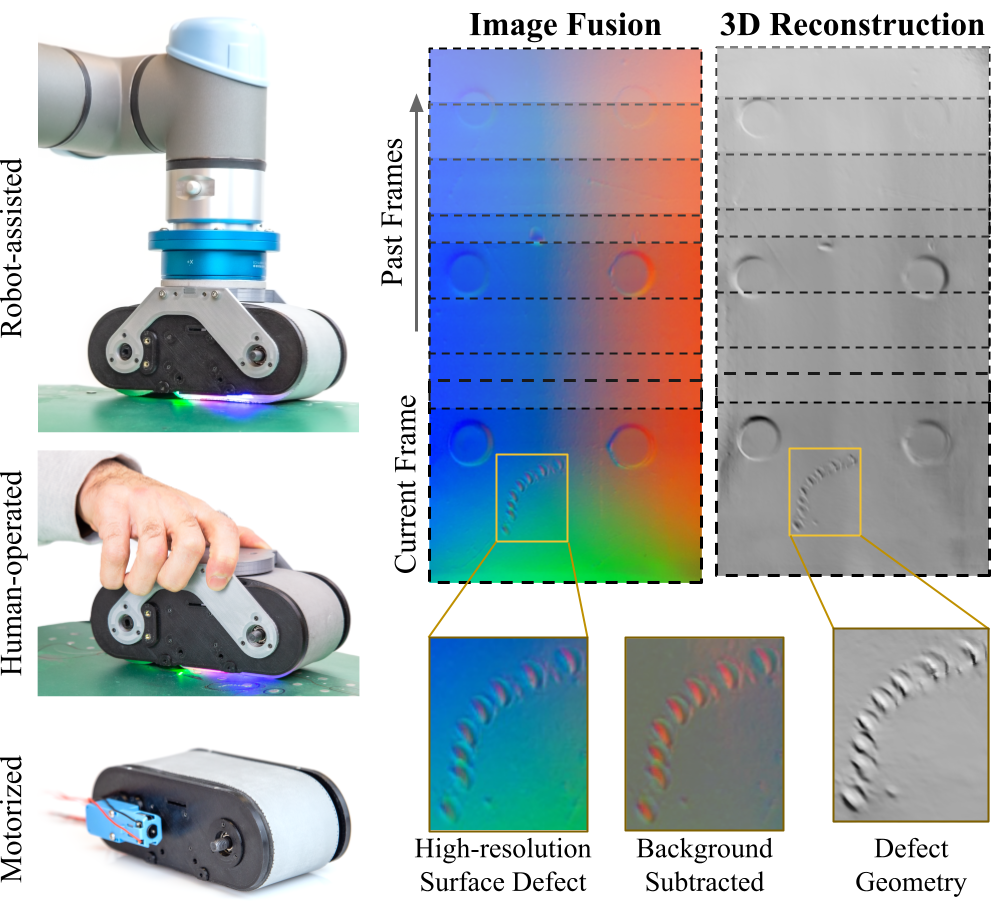}
    \caption{Our GelBelt sensor is mounted on a UR5e robot for continuous surface reconstruction of an aircraft part to detect defects. Gelbelt can potentially be motorized to scan the surface on its own.}
    \label{fig:teaser}
    \vspace{-5 pt}
\end{figure}

Various systems have been developed for sensing detailed information on the surface. For resolutions down to 0.1 mm, systems based on structured light or laser scanning are commonly employed. However, for capturing finer detail, complicated and expensive techniques are used \cite{johnson2011surface}.

As a low-cost and fast technique, vision-based tactile sensors (VBTS) such as GelSight \cite{yuan2017gelsight} have shown promising performance in detailed sensing of objects by using high-resolution camera sensors. They have been widely used in robotic tasks for dexterous manipulation \cite{calandra2018more,yuan2015measurement,li2014localization} and high-resolution surface geometry reconstruction \cite{johnson2011microgeometry,yuan2017gelsight,dong2017improved}. 
Conventionally, a GelSight sensor incorporates a camera and a clear elastomeric membrane coated with a reflective material. The reflective layer deforms, and the camera captures the light variations through the elastomer. A rigid support is attached to the elastomer to press it onto the target surface, imprinting fine surface details into the reflective membrane. 

Numerous designs for VBTSs have been introduced ~\cite{li2014localization,taylor2021gelslim3,finrayliu2022gelsight,brandenroundsensor,tippur2023gelsight360, tippur2024rainbowsight,zhao2023gelsight}, all of which have a small sensing area with a rigid constraint between the elastomer and the supporting membrane, limiting them to attain only local tactile information. The cylindrical-shaped VBTSs ~\cite{shimonomura2021detection,cao2023touchroller,li2023enhanced,yuan2022robot} enabled continuous sensing through their rolling motion. However, their narrow and depth-variant sensing area makes automated rapid movement on the surface challenging. Besides, a larger sensing area requires a much larger cylinder, which is inefficient.

In this paper, we introduce GelBelt as a new mechanical design idea that transcends the limitations of other sensors. GelBelt continuously scans the surface while maintaining large surface contact in each frame. We achieve this by uncoupling the elastomer and the rigid supporting membrane. The elastomer functions as a belt, rolling over two wheels and generating continuous tactile data. Our methodology for surface normal estimation and stitching consecutive frames for surface 3D mesh reconstruction is demonstrated.
We report GelBelt's surface reconstruction accuracy in experiments by the dot product of the estimated and reference surface normal maps, revealing excellent alignment, with an average dot product of more than 0.97 for both single-frame and global surface reconstruction. Also, the reconstruction accuracy of small surface defects is compared with a commercialized micron-scale VBTS. Results further support the GelBelt's capability to obtain high-resolution surface information.

We add markers to the side of the belt to primarily act as a relative position encoder for frame alignment. We also use the marker displacement field to estimate contact normal force up to 60 N with an estimation error of about 1 N (confidence interval 95\%), and surface angle ranging from -10 to 10 degrees and -3 to 3 degrees for roll and pitch with a maximum mean error of 1.38 and 0.1 degrees, respectively. Contact force and angle provide valuable feedback for future work on closed-loop scanning of more challenging surfaces.

GelBelt's mechanical and optical characteristics for rapid and accurate scanning of large surfaces and its versatility for use in automated, manual, or self-driven setups make it an ideal solution for a wide range of applications. GelBelt is scalable and smaller versions can also be used as hand-held tools for rapid surface scanning in industries.

\section{Related Works}
\subsection{Surface Geometry Reconstruction}
As one of the oldest techniques to measure surface 3D topography, mechanical profilometers have been a reliable tool for high-precision measurements. However, their point-by-point approach is time-consuming and less effective for complex geometries or large surfaces \cite{villarrubia1997algorithms}. Non-contact optical methods like laser scanning and structured light scanning have become more prevalent due to their non-contact approach, allowing for faster data acquisition and broader coverage \cite{salvi2004pattern, sansoni2009state}. These optical methods can achieve resolutions down to 0.1 mm \cite{johnson2011surface}, although they often struggle with highly specular and transparent surfaces and changing environmental lighting \cite{hausler2011limitations}.

For applications requiring finer resolution, techniques such as white light interferometry, confocal microscopy, scanning electron microscopy, and atomic force microscopy (AFM) are preferred \cite{russell2001sem, yang2018review, jonkman2020tutorial}. These methods can achieve sub-micron to nanometer-scale resolution, offering unmatched detail. However, their complexity, longer processing time, and cost are significant drawbacks, limiting their use to specialized applications and making them less practical for large-scale scanning.

\subsection{Vision-based Tactile Sensing}
In contrast to other surface scanning techniques, vision-based tactile sensing offers a robust solution, capable of accurately measuring surface conditions across diverse material types and lighting conditions while being low-cost and easy to operate. The high-resolution GelSight tactile sensors \cite{yuan2017gelsight, johnson2011microgeometry,johnson2011surface} boast spatial resolution down to several micrometers, acquiring detailed information about surface topography. They employ cameras and image processing algorithms to analyze the deformation of a soft, clear material, such as silicone elastomers, pressed against a surface. The surface information, such as texture, is captured in the image using the reflective layer on the elastomer \cite{yuan2017gelsight, dong2017improved}.

Introduced VBTSs have a rigid, transparent supporting plate attached to the silicone, pressing it to the target surface. Elastomer's surface traction with the target object and its constraint to the supporting plate prevent these traditional sensors from sliding on the surface while distorting the image signals and tearing the elastomeric membrane. Therefore, for large-scale surfaces, these sensors should be repeatedly pressed on a small area, lifted, and moved to another location \cite{agarwal2023robotic} making these sensors inefficient for continuous scanning and large surface inspection applications.

\subsection{Roller-based Cylindrical VBTSs}
Recently, researchers have proposed new designs to overcome the limitations of conventional VBTSs in continuous sensing. Shimonomura et al. \cite{shimonomura2021detection} designed a cylindrical VBTS incorporating a single light that continuously rolls around its center axis while maintaining contact with the surface. The goal was to detect hard foreign objects like nylon in soft foods in a grey-scale image. Cao et al. \cite{cao2023touchroller} designed TouchRoller, a cylindrical VBTS, to map the 2D texture of a piece of fabric. The sensor mapped an 8 cm x 11 cm flat surface texture (no 3D reconstruction) in 10 s.
Robot in-hand manipulation of small objects using rolling fingertips was studied in ~\cite{lepert2023hand,yuan2022robot,zhang2024rotip}. Employing high-resolution tactile sensing on a rolling finger provided sufficient continuous tactile feedback for in-hand manipulation and reconstruction of small surface geometries.

Large surface reconstruction using cylindrical sensors is challenging as the tactile information corresponds to varying depth levels. Li et al. \cite{li2023enhanced} proposed an image fusion method for cylindrical VBTSs that extracts relevant information associated with various contact depths in the frequency domain and subsequently integrates these distinct characteristics through a differential fusion process. Results suggested the enhanced performance of this method for small indentation compared to the motion distance sampling stitching method.

While recent efforts have explored cylindrical roller design, their optical and mechanical structures exhibit limitations in surface measurement accuracy and speed. The cylinder intersection with a surface produces narrow tactile information in each frame with varying indentation depths, being a big sensing challenge for accurate large surface reconstruction. In contrast, using two wheels in GelBelt provides a larger yet uniform sensing area, preventing missing surface information and enabling accurate rapid scanning of large surfaces.

\section{Sensor Design and Fabrication}

\begin{figure}
    \centering
    \includegraphics[width=\linewidth]{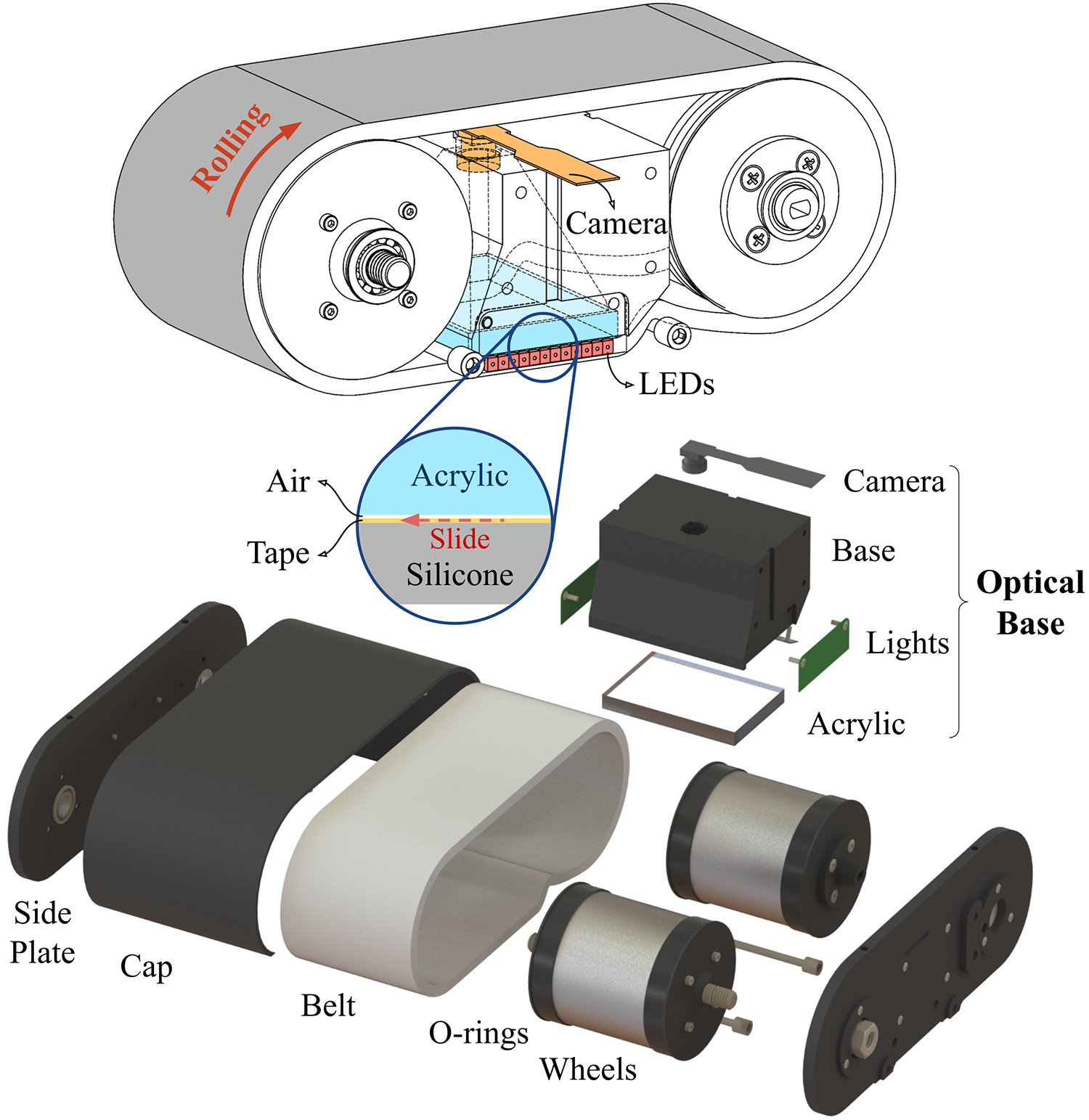}
    \caption{GelBelt's mechanical design. (A) Highlighted optical components. (B) CAD exploded view showing all the components in the model.}
    \label{fig:cad}
    \vspace{-5 pt}
\end{figure}

\subsection{Mechanical Design}
\label{section:design}
We designed GelBelt as a two-wheel structure with a belt made of sensing materials, as demonstrated in \cref{fig:cad}. When scanning a surface, the belt rolls over the wheels while other optical components are fixed between the two wheels to sense the contact information in the area. 
We also add two rows of black markers near the edge of the belt to provide visual reference to the frame transformation and contact forces.
This design enables continuous sensing of surfaces while maintaining a large yet uniform contact area.

The sensing region is the area between the two wheels where acrylic and the belt overlap. There should be minimal friction between the layers in the sensing area to facilitate the belt motion. As elastomers stick to a variety of surfaces, including acrylic, we attach a flexible transparent layer to the belt's inner surface, which has low friction with the acrylic and facilitates the belt's motion without impairing the rolling mechanism. The acrylic-elastomer separation using the intermediate layer introduces a thin air gap, whose effect on the sensor's optical performance is discussed later.

The belt's outer surface is coated with reflective powder to enable sensing, and markers are added near the sides. GelBelt should be pressed onto the surface to imprint the surface detail into the reflective layer. Accordingly, we design aluminum wheels applying initial force through their weight. In contrast to the rigid support, the wheels should have adequate friction with the belt; otherwise, the belt and the wheels will not be mechanically coupled, making them slide on each other and hindering the overall motion. To achieve sufficient friction, rubber O-rings are added to the side of the wheels. 

As an additional option to use the sensor as the end-effector of a robot, the sensor can be self-driven on a surface like a small vehicle. We achieve this by adding two small DC motors (TT Motor Gearbox) on the wheels. \Cref{fig:teaser} bottom shows the motorized GelBelt sensor.

\subsection{Optical Design and Simulation}
\label{section:optical}
As described, the mechanical components enabling the motion mainly consist of the belt and the wheel system. Nonetheless, the optical system adds more system complexities and constraints, directly affecting the sensor performance. Traditionally, optical sensors were designed and improved based on a trial-and-error approach during the fabrication stage. That being said, researchers test different configurations in experiments and modify the model repeatedly to get to the final design which takes a considerable amount of time and effort and does not guarantee the desired outcome.

Recently, a physics-based simulation method has been proposed \cite{agarwal2021simulation} to optimize the VBTS design process and generate tactile images before fabrication. We made use of this method to design the GelBelt's optical system. The simulation was implemented in Blender (version 4.1.0) using the available optical parameters for the lights and surface properties \cite{agarwal2021simulation}. 

\begin{figure}
    \centering
    \includegraphics[width=\linewidth]{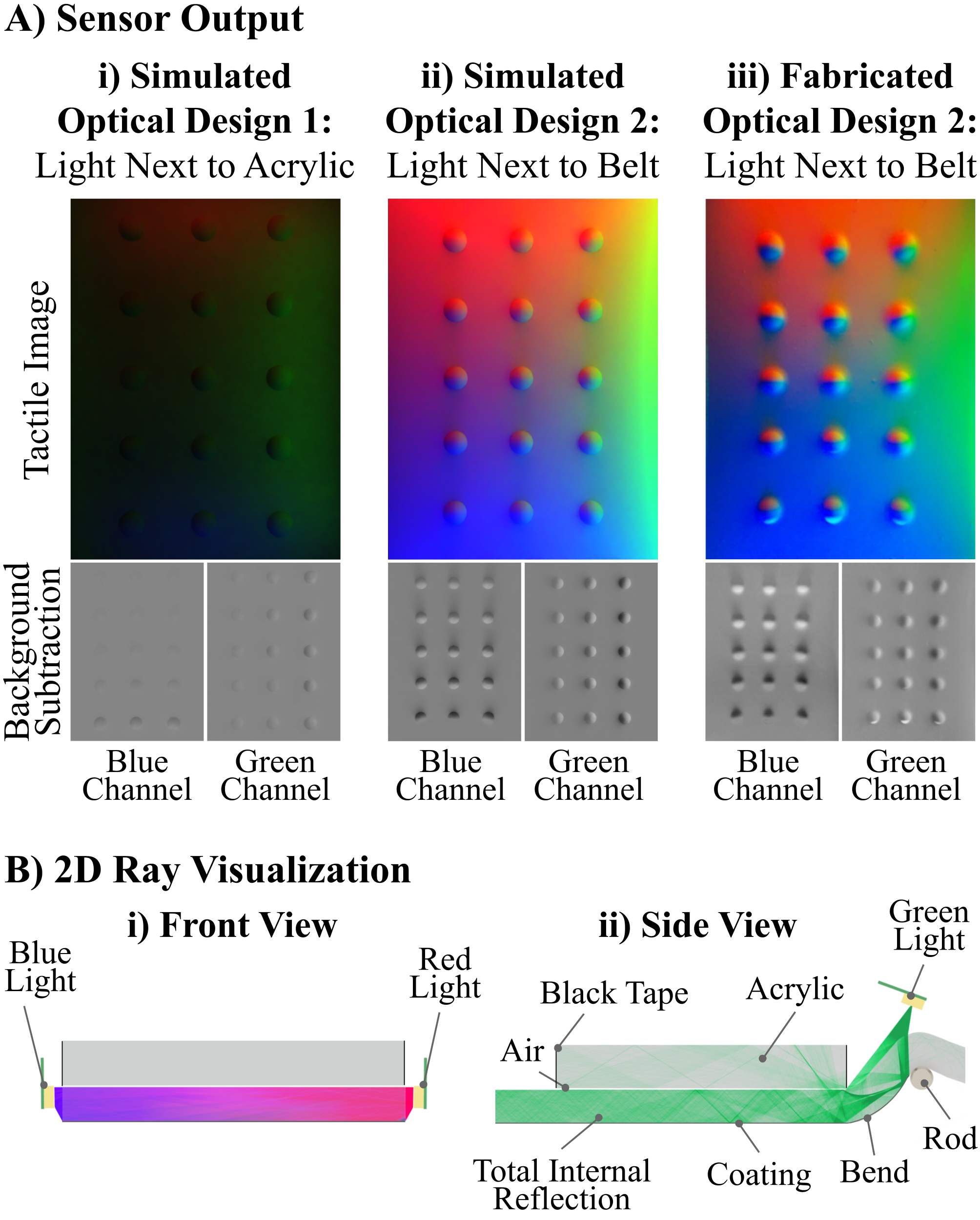}
    \caption{GelBelt's optical configuration. (A) We enhanced the optical system by changing the light locations in simulation to improve the light intensity and contrast over the entire sensing area, which highly matched the real sensor image. (B) A simplified representation of the light placement. The thin air gap between the belt and acrylic results in total internal reflection.}
    \label{fig:optical}
    \vspace{-5 pt}
\end{figure}

Generally, light sources are placed next to the clear acrylic illuminating from the sides of GelSight sensors. Accordingly, we started with the same placements for the light sources in the simulation. This configuration resulted in the tactile image shown in \cref{fig:optical} Ai with multiple spherical indentations. We observed that using the conventional configuration in GelBelt generates dark images with poor information after background subtraction. This issue is caused by the elastomer-acrylic separation and introducing foreknown layers of clear tape and air into the design. consequently, we face total internal reflection (TIR) in the acrylic membrane resulting in poor illumination in terms of intensity and contrast to the sensing surface.

We enhanced the optical design by adjusting the light source locations. The optimal design resulted in the simulated image shown in \cref{fig:optical} Aii with considerably improved lighting. \Cref{fig:optical} Aiii shows the real sensor image which highly matches the simulation results. As shown in \cref{fig:optical} Bi, we placed red and blue light sources on the side of the belt at the contact location. Regarding the green light, we used a small rod with needle bearings to bend the belt next to the acrylic while putting the green light at a specific angle to illuminate the surface. We observed that adding the bend to the belt significantly enhanced the green light illumination over the entire area both in terms of intensity and contrast. \Cref{fig:optical} Bii shows how green rays travel through the elastomer and bounce back from the coating layer into the sensing area.

The GelBelt mechanical design was modified to incorporate the optical system requirements. \Cref{fig:cad} B depicts the exploded CAD view of the GelBelt. Optical components are assembled in a single base for consistent reading in case of disassembling other components. The overall dimensions of the sensor are 175 mm x 80 mm x 65 mm (LxWxH). The sensing area of the sensor measures 40 mm by 60 mm, allowing reconstruction of relatively flat surfaces.

\subsection{Fabrication}
\label{section:fabrication}
We fabricated the belt from Silicone XP-565 (Silinoce Inc). The hardness of the cured silicone depends on the mixing ratio of part A and part B. Accordingly, we made the belt from two layers of silicone that varied in hardness. A soft layer increases sensitivity and a hard layer enhances adhesion to the intermediate layer. The belt is cast in a flat mold (1:7 and 1:16 ratios for hard and soft layers, respectively), and coated with diffusive aluminum powder. To protect the coating, we mixed a 16:1:32 ratio of part A, part B, and Novacs Matte (Smooth-On) and applied it to the surface. We laser engraved two lines of dot markers on belt edges to serve as a position encoder and determine the surface contact condition. Markers had varying intervals to prevent aliasing in displacement measurement.

We used wide clear crystal tape as the intermediate layer. Clear tape is thin and flexible enough to bend over the wheels while having a stiff and slippery surface on the acrylic side. Then, we attached the belt's ends using Sil-Poxy glue (Smooth-On) to form a continuous belt.
The rigid support plate is a rectangle-shaped clear acrylic with a thickness of 6 mm. We designed narrow printed circuit boards for soldering the 3528 SMD LEDs in parallel, matching the light sources used in the simulation step. Consequently, the circuit resistance value for each color was adjusted to match the light intensity in the simulation. Other components were 3D printed with PLA. The fabricated sensor is shown in \cref{fig:teaser}.

\section{Surface Reconstruction}
In this section, We present our large-scale surface reconstruction method using GelBelt. As the belt rolls over textured 3D surfaces, it captures tactile images, which we use to estimate the local surface geometry. Then, we determine the sensor's frame-to-frame planar movement and compose the local geometries into a global 3D shape.

\subsection{Geometry Reconstruction from Single Frame}

GelBelt, like GelSight \cite{yuan2017gelsight} \cite{johnson2011microgeometry} and its derivatives, design hardware for surface normal estimation via the photometric stereo method \cite{johnson2011microgeometry}. We adopt the method from \cite{gelwedge} to predict surface normal maps from Gelbelt's tactile images. Specifically, an 8 mm metal ball is pressed against GelBelt's sensing region at 143 different locations. The contact circles on these tactile images are manually labeled, and the ground truth normal direction within these regions is computed. Using these data, we train a 3-layer MLP (128-32-32) that inputs each pixel's color and coordinates (RGBXY) to predict surface gradients $(g_x, g_y)$. 

During testing, we predict surface gradients for each pixel and convert them into surface normals $\hat{\mathbf{n}}$, forming the surface normal map. Here, $\hat{\mathbf{n}} = \mathbf{n} / \lVert \mathbf{n} \rVert$ and $\mathbf{n} = [g_x, g_y, -1]$.

\subsection{Global Surface Reconstruction}

To get the shape of a larger surface, we need to stitch the measurement from a series of tactile images when the sensor is rolling on the surface. A key challenge in the process is to match the pixels with the real surface location from different frames. We estimate the sensor's planar translational movement between frames by applying optical flow \cite{Lucas1981} to the surface normal map derived from the tactile images. With the initial frame as a reference, the global pose of each frame is obtained by composing these estimated frame-to-frame movements. To estimate the surface normal map of the entire scanned region, we register each local normal map to the global map using the estimated global poses and average the overlapping regions. We use Sigmoid functions on the areas close to the edges as the averaging weights to smoothen the stitching boundaries. Finally, we reconstruct the surface height map of the entire scanned region through Poisson integration \cite{yuan2017gelsight} of the global normal map.

Optical flow works well on tactile images of a surface with non-repeating texture and relatively small motion. To enable sensing with faster motion on repeating textured or non-textured surfaces we use side markers as a position encoder and feed the position as the initial displacement for optical flow. Also, markers are the only useful information in case of sensing a non-textured surface.

Note that our method works for a relatively flat surface. If the surface to scan is of complicated 3D shape, we will need to combine external pose information of the sensor to reconstruct the large-scale 3D shape.

\section{Marker Analysis}

The surface marker motion in VBTSs is the key component for measuring the surface force and torque \cite{yuan2017gelsight,lepora2021soft,mirzaee2023design,gelslim1}. Markers are also essential in robotic manipulation tasks for grasp stability and slip detection \cite{yuan2015measurement,dong2017improved, mirzaee2024multiphysics}.

In Gelbelt, displacement of the side markers can provide useful information such as displacement, force, and contact angle. The cropped regions on the side of the tactile images are considered the marker area, while the rest of the image is used for geometry sensing. Markers have different intervals to prevent mismatching, and we obtain their locations by applying a Blob filter to the red and blue channels of the marker areas next to the corresponding lights. We calculate pixel displacement by matching the marker pattern in two consecutive frames and use the measured displacement in the global surface reconstruction algorithm for coarse alignment of the frames before applying optical flow. 

Contact force and angle are essential feedback in future works on closed-loop control of pressure and orientation of the sensor, improving reconstruction accuracy. We fit two splines to the detected side markers and use the position of 10 points (on each side) with fixed x-axis values as the feature space to train the estimation models. Using the data, we train two MLPs (512-256-128) that input the y values of 20 points. One model outputs the x and y axes rotations, while the other one outputs the normal force. The ground truth force value was measured using a 6-axis F/T sensor (NRS-6050-D80, Nordbo Robotics) on the UR5e robot's wrist.

\begin{figure}
    \centering
    \includegraphics[width=\linewidth]{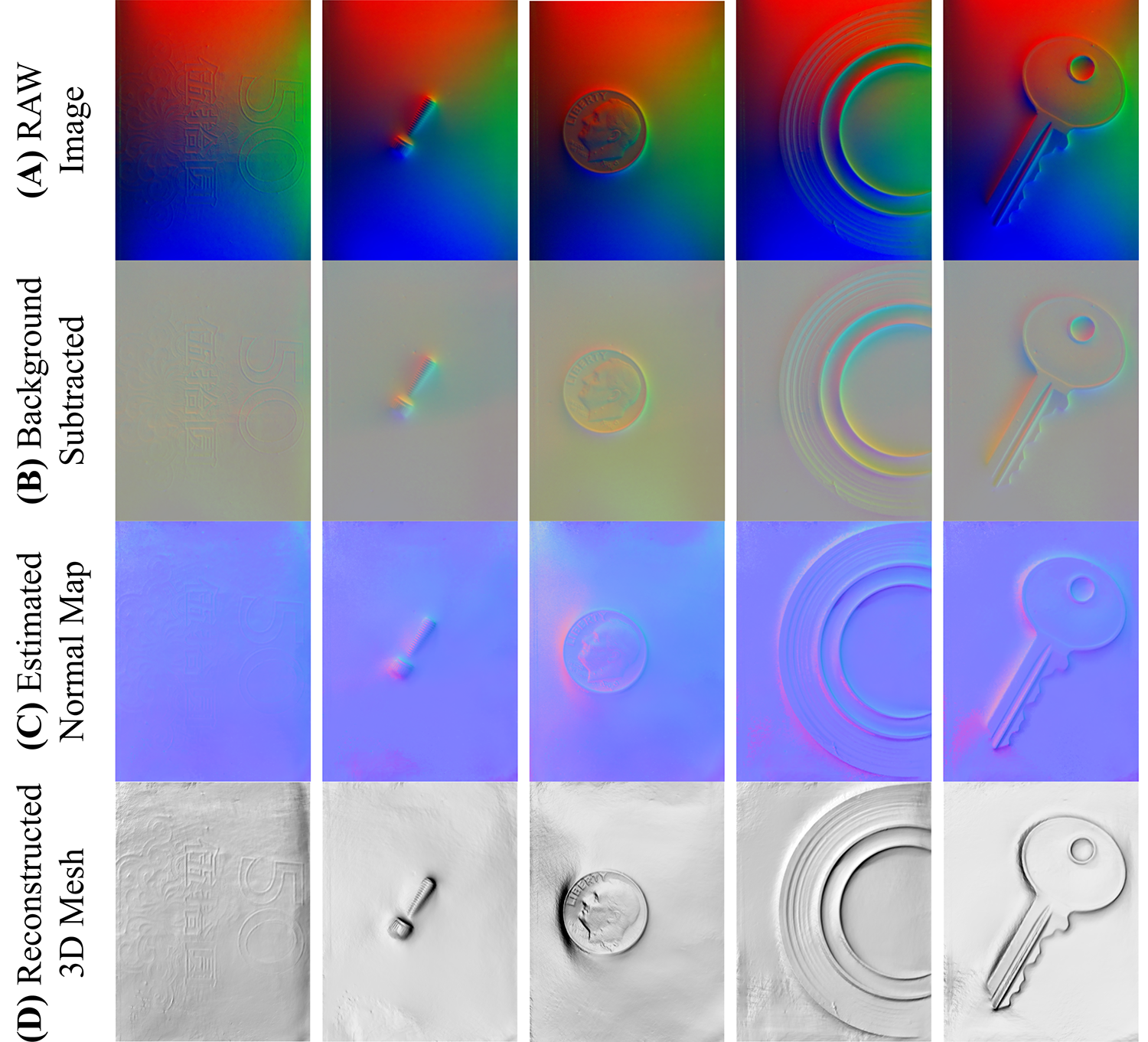}
    \caption{GelBelt's single-frame sensing capability on random objects. From left to right: Chinese Yuan bill, screw, coin, scotch tape, and a key.
    }
    \label{fig:objects}
    \vspace{-5 pt}
\end{figure}

\begin{figure*}
    \centering
    \includegraphics[width=\linewidth]{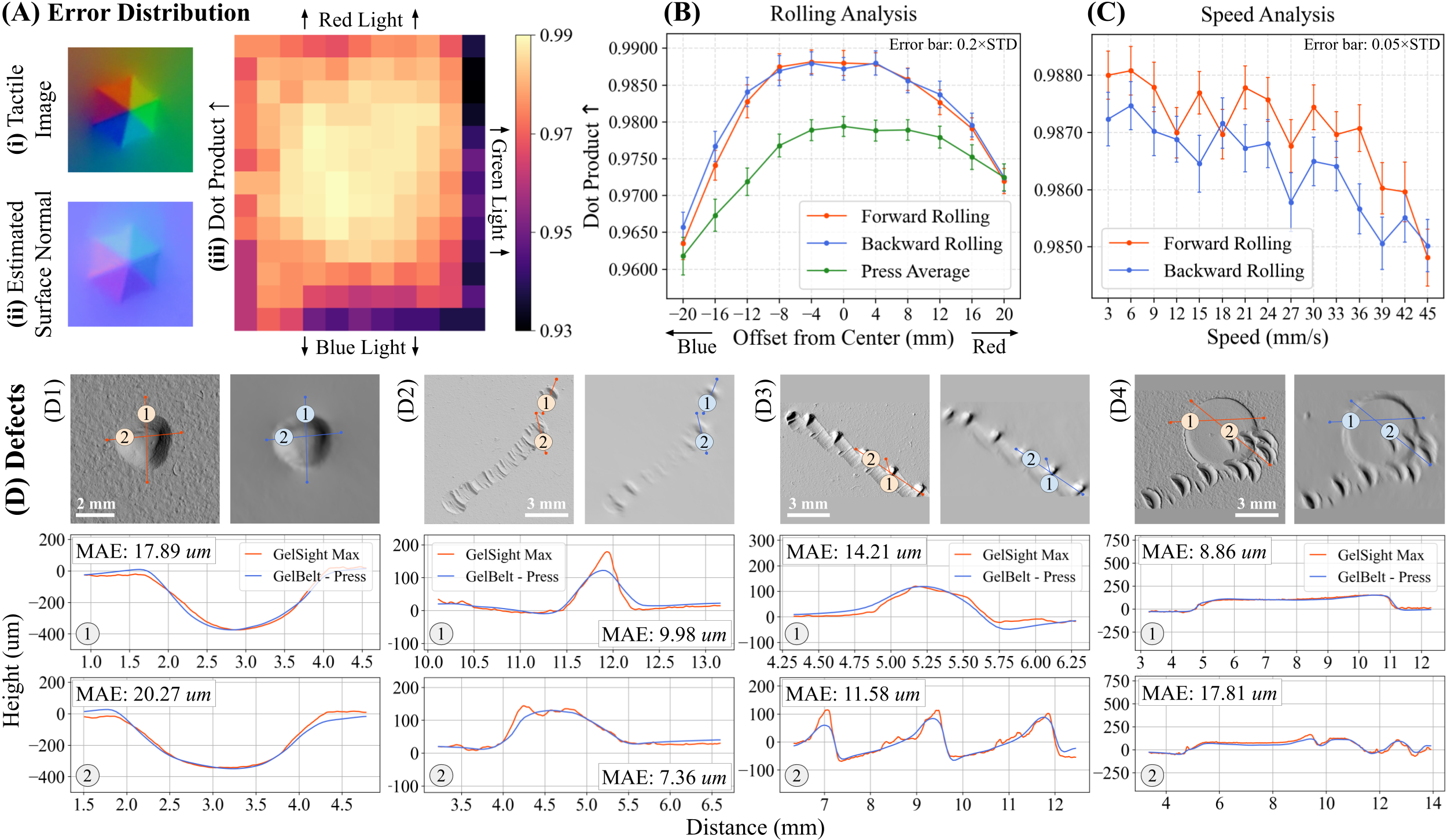}
    \caption{GelBelt reconstruction accuracy. (A) We indent a hex pyramid shape in multiple locations and obtain the 2D normal estimation accuracy plot. (B) We roll over the same object to obtain the accuracy for different offsets from the image centerline. (C) The effect of rolling speed on the sensor accuracy. (D) 3D mesh reconstruction comparison to GelSight Max, a commercialized VBTS with nanometer scale z-axis accuracy.}
    \label{fig:results}
    \vspace{-10 pt}
\end{figure*}

\section{Experiments}

GelBelt can continuously measure surface details and contact conditions. Firstly, we test GelBelt on small objects in a static single frame and evaluate the 3D reconstruction performance by reporting accuracy over the sensing area (\ref{section:experiments:single}). Then, we study large surface reconstruction performance by continuously moving the sensor over planar objects (\ref{section:experiments:large}). Finally, the estimation results for contact angle and force are presented (\ref{section:experiment:marekrs}). For the camera setting, we set a small exposure time, 10 $ms$, to ensure sharpness and minimal motion blur and capped the sampling rate, $\sim$6 to 10 $Hz$, to prevent the collection of redundant frames, although the sampling rate can be further increased for faster motions.

\subsection{Single-frame Reconstruction}
\label{section:experiments:single}
We show the GelBelt's tactile sensing performance by pressing the sensor over objects such as a key, scotch tape, coin, and cash. \Cref{fig:objects} presents the raw images, followed by the corresponding background subtracted image, estimated surface normal, and the reconstructed mesh. As observed, GelBet can obtain detailed surface information for a variety of geometries, from fine texture on a Chinese Yuan bill to a relatively large object such as a key.

We calculate the reconstruction accuracy by pressing a hex indenter with known dimensions in 143 different locations over the sensing area (13x11 grid). We estimate the hex surface normal using the calibrated sensor, mask the indentation area, and calculate the average dot product of the masked hex with the ground truth normal. The dot product of the surface normal vectors as the accuracy metric informs us about the alignment of the normal vectors, which directly affects the reconstructed mesh. The dot product values range from -1 to 1 corresponding to the fully-aligned and opposite-direction vectors, respectively.

\Cref{fig:results} Ai shows a sample tactile image of GelBelt when indented with the hexagonal pyramid geometry. \Cref{fig:results} Aiii illustrates the 2D accuracy plot of the 143 indentations. We observe that the units in the center of the frame have the highest estimation accuracy, and as we get closer to the sides, the dot product value decreases. The lowest accuracy corresponds to the light saturated units at the bottom right corner, adjacent to the green and blue lights. 

Camera filters generally detect R, G, B lights centered at approximately 660, 520, and 450 nm \cite{teague2022review}, and the used green and blue LEDs have a much wider wavelength spectrum (20 nm) compared to red light (7 nm). The closer wavelengths of blue and green lights plus the wider spectrum result in mixed signals in BG channels, decreasing sensitivity in the area close to the green and blue light.

Potentially, we can use GelBelt for surface inspection in quality control and maintenance across various industries. To demonstrate the applicability, small defects on aircraft parts, similar to \cite{agarwal2023robotic}, are 3D reconstructed using GelBelt. We compare the GelBelt's reconstructed mesh with the output mesh obtained from a GelSight Max (GelSight Inc.), a commercialized and highly accurate vision-based tactile sensor with a few microns resolution, as the ground truth. Firstly, we use the Iterative Closest Point (ICP) \cite{besl1992method} method to align the 3D mesh of the two sensors. Then, we obtain five different defect surface profiles on aircraft parts using each sensor and compare the results. As shown in \cref{fig:results}, despite the lower resolution of the GelBelt compared to GelSight Max, it can reconstruct the surface geometry of submillimeter dimensions with good accuracy. GelBelt's performance declines for sensing defects with sharp edges such as \cref{fig:results} D2-1, presenting a smoothed surface profile.

\subsection{Global Surface Reconstruction}
\label{section:experiments:large}

\begin{figure*}
    \centering
    \includegraphics[width=\linewidth]{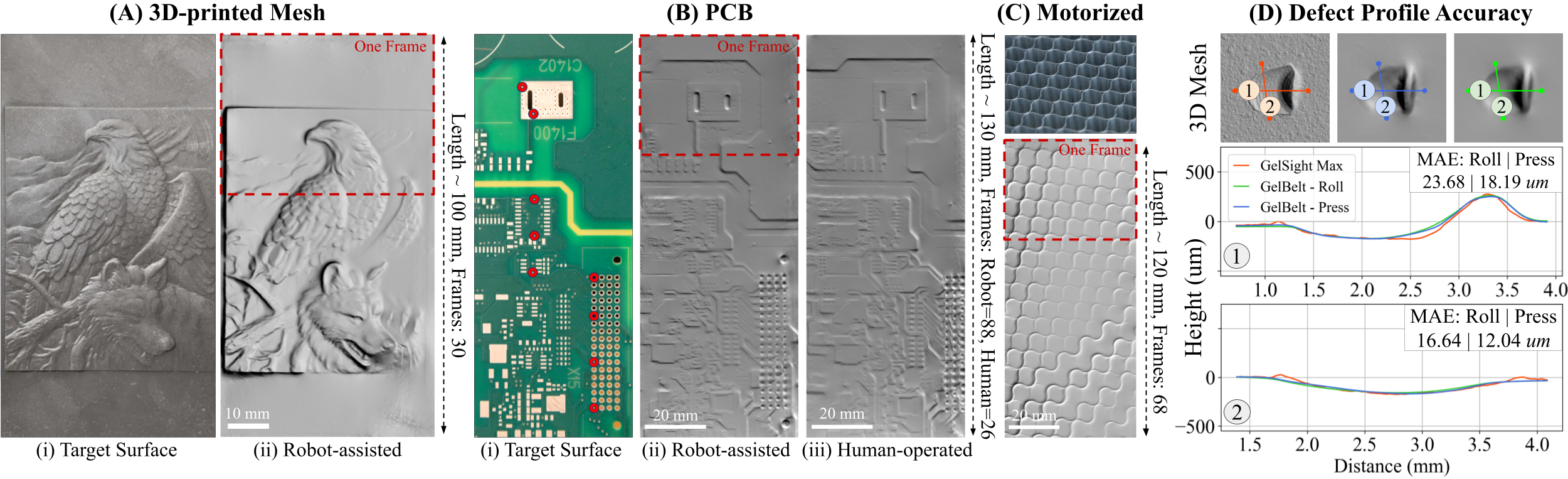}
    \vspace{-18 pt}
    \caption{Large surface reconstruction. (A) Detailed printed mesh surface reconstruction. (B) PCB surface reconstruction in robot-assisted and manual modes. We evaluate the planar distance and angle drift for the red control points. MAE for robot-assisted mode: 0.333 $mm$ and 0.351$^\circ$. For manual mode: 0.381 $mm$ and 0.285$^\circ$. (C) We attached motors for standalone surface sensing. (D) Rolling on the aircraft defects has similar results to single-frame sensing.}
    \label{fig:large}
\end{figure*}

In this section, we study GelBelt's global geometry reconstruction over a larger surface. \Cref{fig:large} Aii and Bii illustrate the reconstructed 3D mesh of a printed mesh, printed using a Form 3+ printer (FormLabs), and a PCB, generated by continuously moving GelBelt over the surface. We observe that GelBelt effectively captures and stitches surface details, allowing for accurate reconstruction of large surfaces. 
Results also highlight the efficiency of our sensor in producing high-quality 3D surface meshes, which is essential in applications requiring rapid yet precise surface scanning.

The global surface reconstruction's error distribution was calculated by rolling GelBelt over the same hexagonal pyramid geometry as the previous section, with different offsets from the center of the image. As shown in \cref{fig:results} B, GelBelt achieves high accuracy with minimal deviations from the reference normal map. Also, similar to the single-frame mode, we observe a higher estimation error close to the blue light. GelBelt's high performance in this section further validates the effectiveness of GelBelt in providing precise surface reconstructions, making it a reliable tool for applications that demand accurate 3D modeling.

\begin{figure*}
    \centering
    \includegraphics[width=\linewidth]{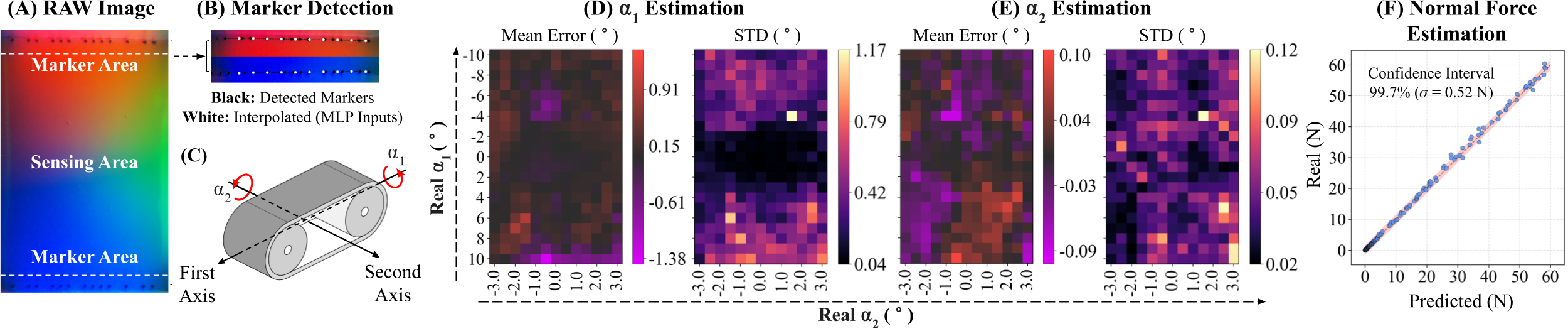}
    \vspace{-18 pt}
    \caption{Contact force and angle estimation. (A) Marker areas on the side of the image. (B) Splines are fit to the detected markers 2x10 points are interpolated as model inputs. (C) Two rotation axes. (D-E) Mean errors and STD plots for each estimated angle. (F) Normal force prediction plot.}
    \label{fig:marker_angel}
    \vspace{-10 pt}
\end{figure*}

Sensor velocity can limit overall accuracy as it controls the number of frames for the same geometry and increases the shear force applied to the sensing surface. We repeated the same procedure as the last experiment with varying speeds, from 3 to 45 $mm/s$, for indentation at the center of the image. Results shown in \cref{fig:results} C, indicate that lower speeds result in higher accuracy of the surface normal estimation while capturing sharper detail of the surface. In contrast, as the scanning speed increases, the error tends to rise with a much smoother estimation of surface detail, suggesting a trade-off between speed and accuracy. GelBelt's scanning rate is substantially higher than the maximum reported values for cylindrical sensors, 11 $mm/s$ \cite{cao2023touchroller}. To the best of our knowledge, GelBelt is the fastest yet accurate VBTS scanning system. A higher frame rate and softer silicone can significantly increase the scanning speed.

So far, GelBelt's movement has been carried out using a UR5e robot. In the next step, a human operator manually moved the sensor over the surface instead of a robot. This test aims to assess GelBelt's performance under more variable and less controlled conditions, simulating a different real-world usage scenario. The results, shown in \cref{fig:large} Biii, demonstrated that despite the inherent inconsistencies introduced by manual operation, Gelbelt captured high-resolution surface information that visually aligns with the results achieved using the robot. We compared the measured distance and angle of all line segments made by 9 control points, red circles in \cref{fig:large} Bi, between the real image (pixel-to-mm scaling obtained using a caliper) and reconstructed PCB surfaces to evaluate the planar drift in the rolling process. The distance and angle mean absolute errors for robot-assisted mode are 0.333 $mm$ and 0.351$^\circ$. For manual mode, the errors are 0.381 $mm$ and 0.285$^\circ$.

Finally, \cref{fig:large} C shows the results of motorized rolling on a honeycomb laser bed. Surface reconstruction results show some inconsistencies and insensitivity in the scanning process. Further refinement is necessary to achieve the same level of accuracy as the manual or robot-assisted modes.

\subsection{Contact Force and Angle}
\label{section:experiment:marekrs}

Markers are detected using the $blob\_dog$ function from the Scikit-image Python library. \Cref{fig:marker_angel} B shows a sample image of the marker area with detected markers (black dots), fitted spline, and feature points (white dots). We collected data for rotations in two axes, shown in \cref{fig:marker_angel} C, using a UR5e robot. Specifically, we mount the sensor on the robot's end effector and change contact angles in the range of -3 to 3 degrees in the x-axis direction (wheelbase axis) and -10 to 10 degrees in the y-axis (wheel axis) with intervals of 0.5 and 1 degree, respectively. We iterate the contact force and angle data acquisition 35 times. After each cycle, GelBelt rolls to get different marker locations. Data was divided into the train, validation, and test sets (60:20:20). 

As shown in \cref{fig:marker_angel} D-E, the model can estimate the surface contact angle of both axes with good accuracy. The estimation error of the first axis is higher compared to the second axis, while larger angles correspond to larger errors. \Cref{fig:marker_angel} F shows that the force estimation model can predict the applied force with good accuracy over the entire range of the study with error of 1 N (95\% confidence interval). Contact force and angle estimation results show the potential application of the markers' motion as feedback in future work on robust scanning of larger surfaces.

\section{Conclusion}

We introduced a new design approach for VBTSs that enables effective rapid scanning of the surface. We designed, fabricated, and tested the proposed sensor, GelBelt, revealing its applicability. We demonstrated GelBelt's mechanical design incorporating an elastomeric belt and two wheels, enabling continuous motion on the surface.

GelBelt can capture tiny details on slightly curved surfaces, reconstruct the surface normal map and 3D mesh, and stitch them together. The qualitative and quantitative analyses highlight the sensor's reliability and precision.
Additionally, GelBelt's reconstructed surface mesh of five small defects was compared with that of a GelSight Max, supporting the accuracy and precision of GelBelt. Results also suggest that reconstruction dot product accuracy was maintained above 0.97 for scanning speeds up to 45 mm/s. Future work will focus on scanning curved surfaces and implementing contact force and angle feedback control.

\addtolength{\textheight}{-0cm}   
\bibliographystyle{IEEEtran}
\bibliography{IEEEabrv, ref}

\end{document}